# Spatial Information Integration in Small Language Models for Document Layout Generation and Classification

Pablo Melendez Abarca
Salzburg University of Applied Sciences
Salzburg, Austria
pablo.melendezabarca@fh-salzburg.ac.at

Clemens Havas
Salzburg University of Applied Sciences
Salzburg, Austria
clemens.havas@fh-salzburg.ac.at

## Abstract
Document layout understanding is a field of study that analyzes the spatial arrangement of information in a document hoping to understand its structure and layout. Models such as LayoutLM (and its subsequent iterations) can understand semi-structured documents with SotA results; however, the lack of open semi-structured data is a limitation in itself. While semi-structured data is common in everyday life (balance sheets, purchase orders, receipts), there is a lack of public datasets for training machine learning models for this type of document. In this investigation we propose a method to generate new, synthetic, layout information that can help overcoming this data shortage. According to our results, the proposed method performs better than LayoutTransformer, another popular layout generation method. We also show that, in some scenarios, text classification can improve when supported by bounding box information.

## 1 Introduction
In the deep learning era, data quantity and quality have become a necessity to keep improving existing models. Different fields benefit from data augmentation techniques, for example, in text classification tasks, changes at character, word, or sentence level are useful [5]. For images, flipping, cropping, or rotating are basic augmentation techniques [27], and the same goes for sound perturbation for audio tasks [1]; however, classical data augmentation techniques [36] can fall short for data-hungry machine learning solutions, since these are limited to creating new samples by modifying existing ones, and manually creating large amounts of new samples is costly and time consuming.

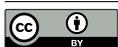

Although generative artificial intelligence has existed for a while, it was not until the second decade of the 21st century that architectures such as the variational autoencoder (VAE) [22] or the generative adversarial network (GAN) [15] were able to output new, artificially generated, data. Later, the transformer architecture was introduced [42], which has allowed for more groundbreaking advancements in the field, the main example being the Generative Pretrained Transformer (GPT) [35] and its subsequent iterations.

Thanks to these advancements, the creation and consumption of new -synthetic- data is now available for fields such as text, audio, images, and sometimes all at the same time multimodal models [6][13][7][33]. However, some specific fields that could benefit from generative solutions for the creation of new data have not been the center of attention. An example of this are semi-structured documents (e.g., receipts, purchase orders, invoices, etc.). Semi-structured documents might have a fixed set of sections that should be present but not necessarily a fixed layout.

Availability for this type of data is scarce, since it usually contains sensitive information like full names, addresses, bank account details, and/or sums of money. This makes downstream tasks (e.g., optical character recognition (OCR) [8][32], intelligent character recognition (ICR) [34]) more challenging since there are not enough public data out there to better train automatic document processing models (e.g., LayoutLM [43]). To solve this challenge, solutions such as LayoutTransformer [16] and LayoutGPT [14] have been proposed to generate new layout samples.

Large language models (LLMs), and particularly small large language models (SLMs), have seen a rise in popularity in the last year, with models such as Llama2-7B [29], Llama3-8B [30] and Mistral-7B [31] leading the charge. Though not as powerful as bigger models, SLMs provide a major advantage which is the possibility of running them locally. Running these models locally has further implications, for instance, increased privacy and reduced costs. Considering these facts, a new layout generation technique is proposed in this work, one that leverages scarce data, as well as the generalization capabilities and increased privacy of SLMs.

Although this research does not create complete synthetic documents, layout generation is a small but important step on the way to generating fully-fledged documents. Generating document layouts with SLMs could also mean that other document processing tasks could be carried out by leveraging SLMs and spatial information, an example of that is document text classification. Reliable, automatic text classification can be crucial when the amount of data at hand is large, as it can reduce processing time and potentially lower the cost [17]. Motivated by that, the second part of the investigation aims to test the efficacy of SLMs for document text classification taking into account the spatial information (i.e., where are the strings in the page?).

A small collection of receipt documents is used in this investigation, in which eight sections of interest have been identified. The header is usually located at the top of the page and might contain the business name and the title of the receipt, if any. The logo is usually at the top of the page and reinforces the brand. The contact section can be in different locations of the receipt (usually at the top) and contains information about the sender or the recipient. The invoice details have been identified all along the left side of the page, and contain information such as receipt number or date of issue. The item(s) table lists the services or products provided and usually stands in the center of the page. The added tax table lists taxes and totals and it is usually placed just after the item(s) table. The payment information section has information about payment (e.g., bank account, payment methods, etc.). Lastly, the footer sits at the bottom and can contain a copyright notice or a 'thank you' message.

This work intends to answer whether, and if so "how", SLMs can be used for the generation of new semi-structured layout data points, and whether spatial information can be used to improve performance in document text classification with language models.

## 2 Related Work

### 2.1 Semi-structured documents

Semi-structured data is characterized for not being completely raw but also for not following a strict, fixed, structure; however, there is not a precise definition of what represents a semi-structured document [2]. Smith et al.[26] suggests that semi-structured documents are defined by a mix of physical structure (e.g., section boundaries) and content indicators (e.g., words in sections). Physical structure elements can be paragraph boundaries, while content indicators can be specific words or titles. In this investigation a small dataset of semi-structured documents, specifically receipts, is used for the experiments.

### 2.2 Document processing: analysis and understanding

The retrieval of information from documents is essential to increase the amount of knowledge [9]. Knowledge does not reside only in the text but also in the layout, Tang et al. [39] define document processing as a two-part process: document analysis and document understanding. Document analysis studies the geometric distribution of the information, while document understanding studies the logical structure of the geometric blocks (spatial information). Hoping to accelerate the process, automatic document processing methods have been proposed, among those are OCR [32][8] and solutions that build upon OCR while including machine learning algorithms and architectures to improve performance (e.g., LayoutLM [43]).

In the case of semi-structured documents, and since machine learning models require large amounts of data to better understand the inputs that produce an output, availability of this type of data (specifically receipts and/or proofs of purchase) becomes essential to improve the performance of these models. Despite the existence of some datasets for document layout analysis most of these belong to other areas, examples of this are PubLayNet [44] and DocBank [24], datasets that contain layout information for scientific articles. Some datasets containing receipt and proof of purchase information exist; however, these are few and are usually quite small, such is the case of FUNSD [19], a dataset of 199 fully-annotated forms. Since receipts contain sensitive information, this might be a reason why there are not many datasets available.

### 2.3 Layout Generation

Using generative AI, different solutions have been proposed to generate layout information for different applications. LayoutGAN [23], as its name implies, makes use of generative adversarial networks [15] to generate new layout information. It is trained on around 25 000 documents containing sections such as heading, title, and captions. The generator takes a set of labels representing graphic elements with random probabilities and samples from both Uniform and Gaussian distributions [23]. A second proposed method is LayoutVAE, based on a variational autoencoder it generates stochastic layouts for different scenes as explained in [21]. LayoutVAE was trained with different datasets for different purposes (none for document layout generation), 5000 training images from the MNIST [11] dataset, and around 113,000 from the COCO dataset [25]. Next comes LayoutTransformer [16], this method improves upon the other two by taking advantage of self-attention to understand the relationships between elements in a scene; for document generation, LayoutTransformer was trained on around 320,000 layouts. To compare the results of our method, we use LayoutTransformer as a baseline.

This work proposes a method that uses local SLMs to produce new document layouts. It is believed that these can perform better thanks to their generalization capabilities (due to the massive amount of pretraining data), and their versatility for automated content generation. Three main advantages have been identified: 1) the use of natural language to specify which elements (labels) should be present in the new layout, 2) the possibility of obtaining accurate results even with less than one hundred documents for fine-tuning thanks to the mix of high-quality training samples and the LLM generalization capabilities, and 3) the ability to create new synthetic layout coordinates. This spatial information can later be used in downstream tasks (e.g: developing fully-fledged documents that can be used to train smaller models for document processing). By using open-source SLMs it is possible to keep the privacy of the data if the application requires it.

### 2.4 Text classification and the impact of spatial information

Using bounding boxes to support document text classification tasks using language models appears to be an unexplored field; nonetheless, [38] states that by explicitly incorporating spatial information into a recurrent neural network for vision applications, classification performance improves, and previously, [18] used two encoding techniques from bag-of-visual-words to add spatial information to text categorization, which resulted in improved results. In the second part of this investigation and to understand if text classification performance improves, spatial information in the form of bounding boxes is added explicitly to the instruction of different SLMs.

## 3 Materials and Methodology

### 3.1 Dataset

The dataset is composed of one hundred and seven (107) semi-structured document images of proofs of purchase. For each of the images, the following information is available:

a) Manually annotated bounding boxes and labels for each section (layout information). Eight elements were identified (Logo: 173, Header: 211, VAT_Table_Summary: 520, PaymentInformation: 558, LineItemTable: 1176, Footer: 450, Contact: 1055, and InvoiceDetails: 1588, for a total of 5731 labeled elements). The software used for manual annotation was LabelStudio [41], an open-source platform for data labeling.

b) All strings in the document and bounding boxes for each of them. Strings and corresponding bounding boxes were obtained with a proprietary OCR solution belonging to the local company.

### 3.2 Language Models

For this investigation, small but capable language models are chosen. Special attention is given to models with good reasoning and text classification capabilities. Models like Meta's Llama2-7B and Mistral's Mistral-7B were considered, but during the development of the investigation better, more capable models were released. These newer models are also small in terms of parameters, but perform much better in many benchmarks. Models are run locally using LM Studio

*3.2.1 Meta's Llama3-8B and Llama3.1-8B.* Meta's Llama3-8B and Llama3.1-8B both present a decoder-only architecture, a tokenizer with a vocabulary of 128K tokens, and grouped query attention (GQA) for better inference. Both models are pretrained on ≈ 15T tokens collected from public sources, with 5% of the dataset in other languages other than English. While Llama3.1-8B is an upgraded version of Llama3-8B, specifics are not provided other than being trained on more, higher quality, synthetic data, having extended context length, stronger reasoning capabilities, and being fully multilingual.

*3.2.2 Google's Gemma2-9B and BERT-base-cased (German).* Open-source Google's model, Gemma2-9B[10], uses a tokenizer with a vocabulary of 256k tokens and GQA, same as Llama3-8B and Llama3.1-8B. The model is pretrained on 8T tokens, has a context length of 8K tokens, and it uses a mix of local sliding window and global attention. The primary language of the pretraining data is English.

Also developed by Google, the Bi-directional Encoder Representations from Transformers model (BERT)[12] for the German language is an open-source model with an encoder-only architecture that is mainly used for question-answering, text generation, summarization, and text classification. It is trained on German Wikipedia, open legal information and news articles.

### 3.3 Llama3-8B for layout generation

*3.3.1 Baseline.* As baseline, the LayoutTransformer model is used with pretrained weights (pretrained for 10 epochs at a learning rate of 1e-5 on 10,000 samples of the PubLayNet dataset for layout generation) and is then fine-tuned with 87% of the small dataset of proofs of purchase for another 40 epochs at a learning rate of 1e-5, using the Adam optimizer. This approach was preferred since training the model from scratch with such a small dataset yielded really poor results.

Table 1: Prompt example for generation of layout information

| | |
|---|---|
| Prompt | Provide bounding box coordinates x1, y1, x2, y2 for these sections of a receipt document: **Logo**, **Contact**, **Header**, **InvoiceDetails**, and **Footer** |
| Answer | **Logo**: 40, 3.143, 96.94, 11.00<br>**Contact**: 9.44, 9.037, 33.33, 16.30<br>**Header**: 3.61, 3.14, 98.61, 10.41<br>**InvoiceDetails**: 8.33, 18.86, 92.77, 30.05<br>**Footer**: 9.17, 90.76, 87.22, 100 |

*3.3.2 Proposed approach.* In the proposed approach, the Llama3-8B base model is finetuned with 87% of the small dataset of proofs of purchase. No pretraining with any other layout dataset is performed. The model is loaded in 4-bit quantized form to be able to fine-tuning it locally and the Llama3-8B tokenizer is used for tokenization. For each sample of the dataset (i.e., each document) labels and coordinates of the bounding boxes are presented as a prompt to the model. The prompt instruction is the following: "Provide bounding box coordinates x1, y1, x2, y2 for these sections of a receipt document: <labels>", where "labels" correspond to every label in the sample separated by a comma. The answer has the following format: "Label: x1, y1, x2, y2" for each label, and each is printed in a new line. See Table 1 for an example. All tokenized entries are padded to ensure that all have the same size at training time. We use Low Rank Adaptation (LoRA) for fine-tuning, with a rank of 32, a scaling factor of 64, a dropout of 0.05, no bias, and targeting all linear layers of the model. The model is then fine-tuned for 4 epochs at a learning rate of 1.5e-4 using the AdamW optimizer to optimize learning rate and weight decay separately. Once the model has been finetuned, the next step is to prompt it to return layout information based on provided labels by the user. Finally, the generated coordinates for each label are drawn using Python (as shown in Figure 1).

Since LayoutTransformer is not conditional (i.e., desired labels cannot be provided in advance) an image-to-image comparison is not a good way to measure performance, instead the performance of the models is measured following this process: Multiple layout samples are generated for both models, and bounding boxes for individual labels are collected. For each label, two clusters are created: one that contains all the origin points (top-left corner) and a second one that contains all the closing points (bottom-right corner) of the bounding boxes. Label clusters are also created for the testing subset, which contains the remaining 13% of the layouts, and serves as ground truth. Once all clusters have been created for the baseline and for the proposed approach, cluster centroids and bounding box average area sizes are calculated at label level. These are then compared to those of the ground truth; Mahalanobis distance between centroids and area size differences are provided and used as performance indicators. Mahalanobis distance is preferred over the Euclidean distance since it accounts for correlation between samples [28]. That is to say, the probability distribution of the samples of each label is considered when measuring the distance, since it might happen that a sample appears to be too far away from the centroid but it actually exists within the probability ellipsis for a given label. A number of receipt documents in the dataset contain more than one instance of the same label, for that reason intra-label overlapping is also reported.

### 3.4 Text classification with spatial information

The small dataset of proofs of purchase makes available all strings and their corresponding bounding boxes. Since labels and bounding boxes for each document's main sections are also available, it is of interest to run a text classification experiment to determine the impact of adding spatial information. For this purpose, the data are further prepared: For each document, each string is associated with one of the main section labels, this is done by determining which main section bounding box contains the string's bounding box. If the string is contained by more than one main section, then the section with the smallest bounding box size is chosen as the string's new label. For example, <string> is contained by sections "Contact" and "Header", "Contact"'s area is 100 and "Header"'s area is 200, in this case <string> is classified as "Contact". If a string does not fully fall within a main section bounding box, the string is classified as "Undefined" and removed. In the end, there are 5731 classified strings as ground truth, 85% is used for training and validation, and 15% for testing.

*3.4.1 BERT for text classification.* BERT model is fine-tuned to perform text classification. Two approaches are followed. In the first approach, the model is fine-tuned without adding any type of spatial information, in the second approach, spatial information is explicitly added to the string in the form of coordinates only separated by a single space. In both training sessions, the model is trained for 3 epochs, with a learning rate of 5e-5, using cross-entropy loss function and AdamW optimizer.

*3.4.2 Large language models for text classification.* For text classification, Llama3-8B, Llama3.1-8B and Gemma2-9B are also prompted. In this case, the SLMs are not fine-tuned but a few-shot approach is used when prompting them. This is done to test the generalization capabilities of larger, more capable models. Same as with BERT, in the first run no spatial information is provided, and in the second run bounding box coordinates are added, both for the examples and for the non-classified strings. To avoid any issues with the context window of the models, the testing set is divided into batches of 60 strings with a final batch of 51 strings.

For all models, classification accuracy and the weighted F1, precision and recall scores are provided. We report the weighted scores for F1, precision, and recall due to our dataset not having the same amount of samples for each class.

## 4 Experimental results

### 4.1 Layout generation

As seen in Table 2 LayoutTransformer and the proposed method with Llama3-8B got pretty similar results with respect to the Mahalanobis distance to origin and closing points. The distance to the bounding box origin point for the Contact (C), Logo (L), Header (H), and Footer (F) labels was closer to the ground truth when using LayoutTransformer, while the distance to the origin for the InvoiceDetails (I), LineItemTable (LIT), VAT_Table_Summary (VAT), and PaymentInformation (P) labels was closer to the ground truth when using the proposed method with Llama3-8B. The distance to the bounding box closing point for H, VAT and P labels was closer to the ground truth when using LayoutTransformer, while the distance to the origin point for C, L, I, LIT, and F labels was closer to the ground truth when using the proposed method.

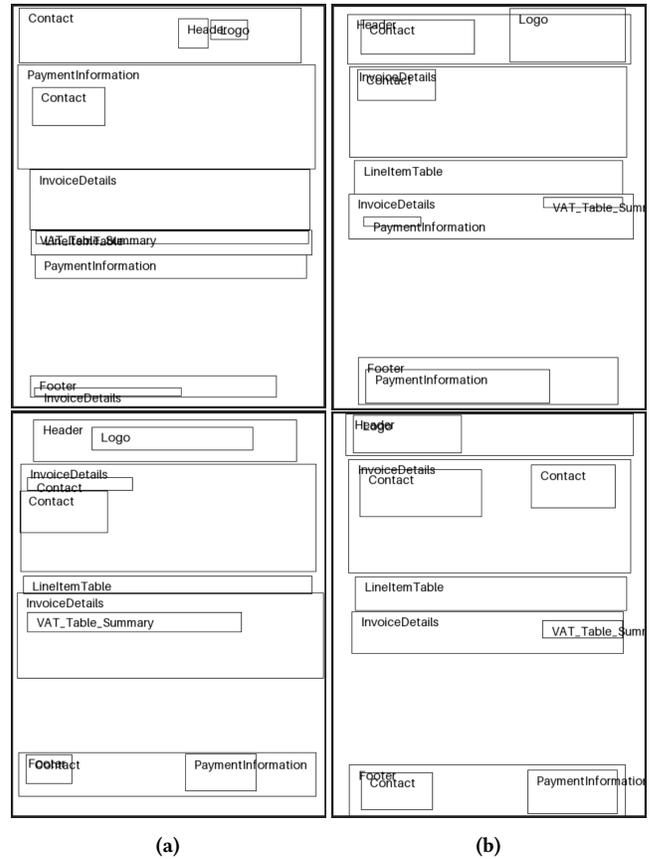

Figure 1: Dataset samples to the left (a) vs Llama3-8B synthetic samples to the right (b) with same labels

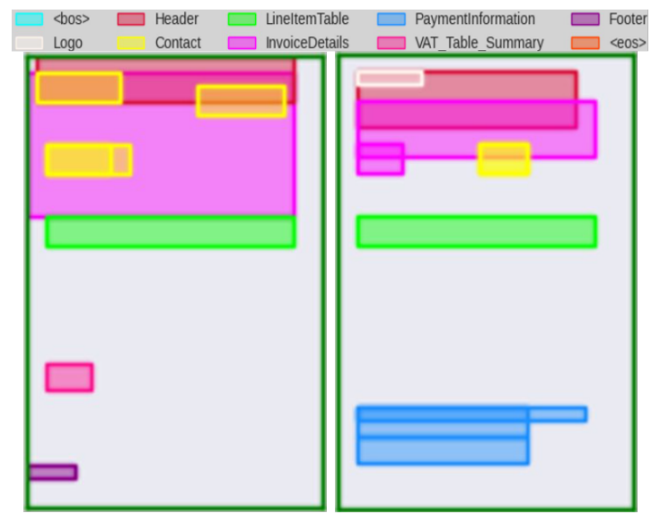

Figure 2: LayoutTransformer samples

Table 2: Comparison between LayoutTransformer and Llama3-8B for layout generation

| Labels | Ground Truth | LayoutTransformer | | | | Llama3-8B for Layout Generation | | | |
|---|---|---|---|---|---|---|---|---|---|
| | BoxArea | Mah. Distance | | Area Diff. | Overlaps | Mah. Distance | | Area Diff. | Overlaps |
| | | Origin | Closing | | | Origin | Closing | | |
| C | 180.90 | **0.27** | 0.42 | +81.32 | 8 | 0.34 | **0.41** | **+66.34** | 1 |
| L | 235.22 | **0.37** | 0.43 | **+22.19** | - | 0.7 | **0.37** | +181.3 | - |
| H | 1145.77 | **0.44** | **0.5** | **-260.6** | 1 | 0.71 | 0.92 | -268.84 | - |
| I | 1571.65 | 0.2 | 0.51 | -866.95 | 6 | **0.19** | **0.32** | **-223.82** | - |
| LIT | 801.76 | 0.77 | 1.13 | -310.65 | 1 | **0.67** | **0.65** | **-236.28** | - |
| VAT | 259.53 | 1.07 | **0.22** | +73.8 | - | **0.5** | 0.38 | **+31.21** | - |
| P | 241.31 | 0.73 | **0.41** | -38.75 | 5 | **0.5** | 0.79 | **-36.49** | - |
| F | 901.72 | **0.24** | 0.78 | -519.5 | 2 | 0.62 | **0.77** | **-53.69** | - |

A bigger difference was noticed between the two methods with respect to the average bounding box area size. Only for labels L and H was LayoutTransformer closer to the ground truth, and in the case of H the difference was particularly small (LayoutTransformer average area difference was -260.6 units, while the area difference for the proposed method was -268.84 units). For the other six labels, the proposed method outperformed LayoutTransformer, mainly for labels I and F, with differences that exceeded the 400 units.

The biggest difference was observed when comparing intralabel bounding box overlapping. Using LayoutTransformer, eight overlaps were found for C, one for H, six for I, one for LIT, five for P, and two for F (see Figure 2); however, using the proposed method, we only reported one overlapping bounding box for the C label and none for the rest. In this sense, the proposed method with Llama3-8B clearly outperformed generation with LayoutTransformer.

### 4.2 Text classification

For each SLM, the classification task was performed three different times for each scenario (without bounding boxes vs. with bounding boxes). With respect to text classification (see Table 3) with non-finetuned SLMs without bounding boxes in the prompt, it was observed that, as expected, models with more parameters performed better, as seen by Gemma2-9B achieving 47% accuracy with a standard deviation of 0.58%, then followed by the most recent version of Meta's Llama at the moment of writing (Llama3.1-8B) with 44% accuracy with a standard deviation of 0.58%. The original Llama3-8B achieved the lowest accuracy with 42% with a standard deviation of 3.21%. Nonetheless, when doing text classification with non-finetuned SLMs using bounding boxes within the prompts, no significant difference in accuracy was observed for Gemma2-9B, which remained at 47% with a standard deviation of 1.15%. Llama3.1-8B worsened with an accuracy of 44% with a standard deviation of 2.08% and Llama3-8B also went down by 1% with a standard deviation of 2.89%.

Despite the massive parameter difference between BERT and the other SLMs, when using finetuned BERT without bounding boxes the text classification accuracy went up to 62% with a standard deviation of 4.51%, outperforming all other SLMs by at least 15%, and unlike the SLMs, finetuning BERT concatenating the bounding box information increased the text classification accuracy by 12% with a standard deviation of 4.13% (thus outperforming the other SLMs by at least 27%).

## 5 Discussion

Results obtained for layout generation using the proposed method with Llama3-8B showed that this approach is at least as good as LayoutTransformer when it comes to positioning of the origin ($x_1$, $y_1$) and closing ($x_2$, $y_2$) points of the bounding boxes. Nonetheless, the similarity between both methods in this regard might be explained by the multiple intralabel overlapping (see Table 2) when using LayoutTransformer, which might give this method an unfair advantage since complete overlapping (e.g., two or more bounding boxes having the exact same coordinates) was noticed in some of the instances, thus reducing the Mahalanobis distance.

Regarding the area size of the bounding boxes, the proposed method performed better, not only by better approaching the ground truth area sizes of most labels, but also by respecting interlabel dimensions, as shown in Figure 3. As an example, within the ground truth samples, the average area size of label I exceeded that of label H, which is also the case for the new generations using Llama3-8B, in the case of LayoutGeneration, the average area size of label H exceeded that of label I. The same behavior could be observed between labels LIT and F.

For text classification, it was expected to see BERT perform better by adding spatial information at training time; however, an improvement of 12% in accuracy came as a surprise given the simplicity of the approach (i.e., simply concatenating spatial information to the strings). On the other hand, the task was more challenging for the SLMs given the simpler approach (few-shot prompting); however, spatial information having no impact in accuracy also came unexpectedly. Though adding bounding boxes did not have any impact in accuracy for text classification, different observations were made; for example, classification precision was particularly low for L, H and F labels, all of which are present mostly at the top or the bottom of the page which suggests that the models are struggling to understand these sections of the document; however, it is important to consider other potential causes, such as bad quality samples, the low amount of instances for the aforementioned labels, and lastly, the bi-directional nature of BERT, which lets it gain context from left and right unlike auto-regressive models.

Table 3: Text classification with BERT

| Model | No Bounding Boxes | | | | Bounding Boxes | | | |
|---|---|---|---|---|---|---|---|---|
| | Acc. | F1 | Prec. | Rec. | Acc. | F1 | Prec. | Rec. |
| BERT (FT) | **0.62** | 0.51 | 0.60 | 0.50 | **0.74** | 0.67 | 0.71 | 0.65 |
| Llama3-8B | 0.42 | 0.41 | 0.45 | 0.46 | 0.42 | 0.41 | 0.45 | 0.45 |
| Llama3.1-8B | 0.44 | 0.41 | 0.44 | 0.43 | 0.42 | 0.38 | 0.45 | 0.41 |
| Gemma2-9B | 0.47 | 0.46 | 0.5 | 0.46 | 0.47 | 0.46 | 0.49 | 0.46 |

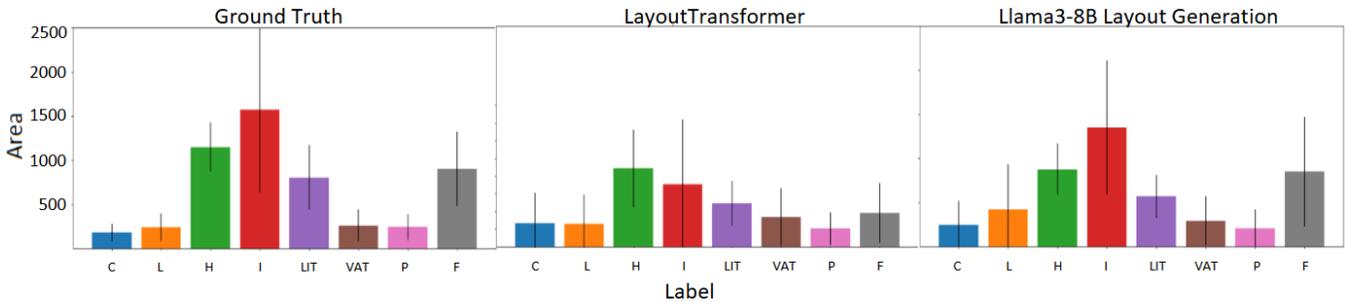

Figure 3: Labels by bounding box total area

In all iterations of the classification task and when using Llama3-8B (the least capable of the three tested SLMs), it was also observed that not adding bounding boxes increased issues such as the amount of missed and merged strings, as well as lower string faithfulness. This behavior was not observed with Llama3.1-8B or Gemma2-9B.

Finally, it was also noticed that all the tested models had a much lower classification accuracy for strings labeled as L, H, or F, with the peculiarity that most of these segments are located at the top and bottom of the page.

## 6 Conclusion

In this investigation, an LLM-powered, layout generation technique has been provided by finetuning a Llama3-8B model using only a limited amount of labeled bounding boxes. The results have shown that the proposed method outperforms the LayoutTransformer approach (which additionally had to be pretrained on a much bigger dataset), especially when multiple instances of the same label exist and intra-label overlapping is undesired. By taking advantage of LLM prompting, the proposed method is also fully conditional, meaning that the user is in charge of specifying which and how many labels are wanted in the page. Although the initial results for all labels are largely positive, generalizing these findings across all labels may be premature, given the varying number of instances between them, which range into the hundreds. Underrepresented labels, such as L and H, would benefit from more instances to further influence the weights of the network.

In the future, it would be valuable to compare this feature more extensively with other conditional methods. New, synthetic, layout samples could be later used for other downstream tasks, such as continuing to create fully synthetic proof of purchase documents or even feeding other models for further training. The later must be done carefully, since it has been demonstrated that models decrease quality when trained on recursively generated data [4][37]; however, Meta and Google have shown that there is a right way to do knowledge transfer via knowledge distillation [40]. This work also showed that accuracy improves by simply concatenating spatial information to the target strings when fine-tuning BERT for text classification; however, this cannot be said of a simpler method like few-shot learning even if using bigger, more capable, models such as Llama3.1-8B and Gemma2-9B. In the future, and since context window is no longer a limitation for recent models, a many-shot approach [3][20] where many examples are provided at prompting time and that better represents each label's ratio might also yield better results. Lastly, improving the quality of the samples and limiting the text classification task to just include the most problematic labels (those with lower precision) are also worth testing.

## Acknowledgments

The authors would like to thank the company Blumatix Intelligence GmbH for providing the dataset for the experiments and Rudolf Dittrich (Blumatix Intelligence GmbH) for valuable feedback on our results.

The work is supported by a grant from the Federal State Government of Salzburg, Austria (WISS-FH no. 20204-WISS/140/584/15-2023)

# References

[1] Olusola O. Abayomi-Alli, Robertas Damaševičius, Atika Qazi, Mariam Adedoyin-Olowe, and Sanjay Misra. 2022. Data Augmentation and Deep Learning Methods in Sound Classification: A Systematic Review. *Electronics* 11, 22 (2022). https://doi.org/10.3390/electronics11223795

[2] Serge Abiteboul. 1997. Querying semi-structured data. In *Database Theory — ICDT '97*, Foto Afrati and Phokion Kolaitis (Eds.). Springer Berlin Heidelberg, Berlin, Heidelberg, 1–18.

[3] Rishabh Agarwal, Avi Singh, Lei M Zhang, Bernd Bohnet, Luis Rosias, Stephanie C.Y. Chan, Biao Zhang, Aleksandra Faust, and Hugo Larochelle. 2024. Many-shot In-Context Learning. In *ICML 2024 Workshop on In-Context Learning*. https://openreview.net/forum?id=goi7DFHlqS

[4] Sina Alemohammad, Josue Casco-Rodriguez, Lorenzo Luzi, Ahmed Imtiaz Humayun, Hossein Babaei, Daniel LeJeune, Ali Siahkoohi, and Richard Baraniuk. 2024. Self-Consuming Generative Models Go MAD. In *The Twelfth International Conference on Learning Representations*. https://openreview.net/forum?id=ShjMHfmPs0

[5] Markus Bayer, Marc-André Kaufhold, and Christian Reuter. 2022. A Survey on Data Augmentation for Text Classification. *ACM Comput. Surv.* 55, 7, Article 146 (dec 2022), 39 pages. https://doi.org/10.1145/3544558

[6] Zalán Borsos, Raphaël Marinier, Damien Vincent, Eugene Kharitonov, Olivier Pietquin, Matt Sharifi, Dominik Roblek, Olivier Teboul, David Grangier, Marco Tagliasacchi, and Neil Zeghidour. 2023. AudioLM: A Language Modeling Approach to Audio Generation. *IEEE/ACM Transactions on Audio, Speech, and Language Processing* 31 (2023), 2523–2533. https://doi.org/10.1109/TASLP.2023.3288409

[7] Tom Brown, Benjamin Mann, Nick Ryder, Melanie Subbiah, Jared D Kaplan, Prafulla Dhariwal, Arvind Neelakantan, Pranav Shyam, Girish Sastry, Amanda Askell, Sandhini Agarwal, Ariel Herbert-Voss, Gretchen Krueger, Tom Henighan, Rewon Child, Aditya Ramesh, Daniel Ziegler, Jeffrey Wu, Clemens Winter, Chris Hesse, Mark Chen, Eric Sigler, Mateusz Litwin, Scott Gray, Benjamin Chess, Jack Clark, Christopher Berner, Sam McCandlish, Alec Radford, Ilya Sutskever, and Dario Amodei. 2020. Language Models are Few-Shot Learners. In *Advances in Neural Information Processing Systems*, H. Larochelle, M. Ranzato, R. Hadsell, M.F. Balcan, and H. Lin (Eds.), Vol. 33. Curran Associates, Inc., 1877–1901. https://proceedings.neurips.cc/paper_files/paper/2020/file/1457c0d6bfcb4967418bfb8ac142f64a-Paper.pdf

[8] Arindam Chaudhuri, Krupa Mandaviya, Pratixa Badelia, and Soumya K. Ghosh. 2017. *Optical Character Recognition Systems.* Springer International Publishing, Cham, 9–41. https://doi.org/10.1007/978-3-319-50252-6_2

[9] C. H. Chen. 2005. *Handbook Of Pattern Recognition And Computer Vision.* World Scientific Publishing Co., Inc., USA.

[10] Google DeepMind. 2024. *Gemma 2: Improving Open Language Models at a Practical Size.* Google DeepMind Research Technical Report. Google.

[11] Li Deng. 2012. The MNIST Database of Handwritten Digit Images for Machine Learning Research [Best of the Web]. *IEEE Signal Processing Magazine* 29, 6 (2012), 141–142. https://doi.org/10.1109/MSP.2012.2211477

[12] Jacob Devlin, Ming-Wei Chang, Kenton Lee, and Kristina Toutanova. 2019. BERT: Pre-training of Deep Bidirectional Transformers for Language Understanding. In *North American Chapter of the Association for Computational Linguistics*. https://api.semanticscholar.org/CorpusID:52967399

[13] P. Esser, R. Rombach, and B. Ommer. 2021. Taming Transformers for High-Resolution Image Synthesis. In *2021 IEEE/CVF Conference on Computer Vision and Pattern Recognition (CVPR)*. IEEE Computer Society, Los Alamitos, CA, USA, 12868–12878. https://doi.org/10.1109/CVPR46437.2021.01268

[14] Weixi Feng, Wanrong Zhu, Tsu-Jui Fu, Varun Jampani, Arjun Akula, Xuehai He, S Basu, Xin Eric Wang, and William Yang Wang. 2023. LayoutGPT: Compositional Visual Planning and Generation with Large Language Models. In *Advances in Neural Information Processing Systems*, A. Oh, T. Naumann, A. Globerson, K. Saenko, M. Hardt, and S. Levine (Eds.), Vol. 36. Curran Associates, Inc., 18225–18250. https://proceedings.neurips.cc/paper_files/paper/2023/file/3a7f9e485845dac274223375c934cb4db-Paper-Conference.pdf

[15] Ian Goodfellow, Jean Pouget-Abadie, Mehdi Mirza, Bing Xu, David Warde-Farley, Sherjil Ozair, Aaron Courville, and Yoshua Bengio. 2020. Generative adversarial networks. *Commun. ACM* 63, 11 (oct 2020), 139–144. https://doi.org/10.1145/3422622

[16] Kamal Gupta, Justin Lazarow, Alessandro Achille, Larry S. Davis, Vijay Mahadevan, and Abhinav Shrivastava. 2020. LayoutTransformer: Layout Generation and Completion with Self-attention. *2021 IEEE/CVF International Conference on Computer Vision (ICCV)* (2020), 984–994. https://api.semanticscholar.org/CorpusID:238247253

[17] Emmanouil Ikonomakis, Sotiris Kotsiantis, and V. Tampakas. 2005. Text Classification Using Machine Learning Techniques. *WSEAS transactions on computers* 4 (08 2005), 966–974.

[18] Radu Tudor Ionescu and Marius Popescu. 2016. *Spatial Information in Text Categorization.* Springer International Publishing, Cham, 229–241. https://doi.org/10.1007/978-3-319-30367-3_9

[19] Guillaume Jaume, Hazim Kemal Ekenel, and Jean-Philippe Thiran. 2019. FUNSD: A Dataset for Form Understanding in Noisy Scanned Documents. In *2019 International Conference on Document Analysis and Recognition Workshops (ICDARW)*, Vol. 2. 1–6. https://doi.org/10.1109/ICDARW.2019.10029

[20] Yixing Jiang, Jeremy Andrew Irvin, Ji Hun Wang, Muhammad Ahmed Chaudhry, Jonathan H Chen, and Andrew Y. Ng. 2024. Many-Shot In-Context Learning in Multimodal Foundation Models. In *ICML 2024 Workshop on In-Context Learning*. https://openreview.net/forum?id=j2rKwWXdcz

[21] A. Jyothi, T. Durand, J. He, L. Sigal, and G. Mori. 2019. LayoutVAE: Stochastic Scene Layout Generation From a Label Set. In *2019 IEEE/CVF International Conference on Computer Vision (ICCV)*. IEEE Computer Society, Los Alamitos, CA, USA, 9894–9903. https://doi.org/10.1109/ICCV.2019.00999

[22] Diederik P. Kingma and Max Welling. 2013. Auto-Encoding Variational Bayes. *CoRR* abs/1312.6114 (2013). https://api.semanticscholar.org/CorpusID:216078090

[23] Jianan Li, Jimei Yang, Aaron Hertzmann, Jianming Zhang, and Tingfa Xu. 2019. LayoutGAN: Generating Graphic Layouts with Wireframe Discriminators. In *7th International Conference on Learning Representations, ICLR 2019, New Orleans, LA, USA, May 6-9, 2019*. OpenReview.net. https://openreview.net/forum?id=HJxB5sRcFQ

[24] Minghao Li, Yiheng Xu, Lei Cui, Shaohan Huang, Furu Wei, Zhoujun Li, and Ming Zhou. 2020. DocBank: A Benchmark Dataset for Document Layout Analysis. In *Proceedings of the 28th International Conference on Computational Linguistics*, Donia Scott, Nuria Bel, and Chengqing Zong (Eds.). International Committee on Computational Linguistics, Barcelona, Spain (Online), 949–960. https://doi.org/10.18653/v1/2020.coling-main.82

[25] Tsung-Yi Lin, Michael Maire, Serge Belongie, James Hays, Pietro Perona, Deva Ramanan, Piotr Dollár, and C. Lawrence Zitnick. 2014. Microsoft COCO: Common Objects in Context. In *Computer Vision – ECCV 2014*, David Fleet, Tomas Pajdla, Bernt Schiele, and Tinne Tuytelaars (Eds.). Springer International Publishing, Cham, 740–755.

[26] M Lopez and DJ Smith. 1997. Information extraction for semi-structured documents. Proc. Workshop on Management of Semi-structured Data ; Conference date: 01-01-1997.

[27] Kiran Maharana, Surajit Mondal, and Bhushankumar Nemade. 2022. A review: Data pre-processing and data augmentation techniques. *Global Transitions Proceedings* 3, 1 (2022), 91–99. https://doi.org/10.1016/j.gltp.2022.04.020 International Conference on Intelligent Engineering Approach(ICIEA-2022).

[28] G. Mclachlan. 1999. Mahalanobis Distance. *Resonance* 4 (06 1999), 20–26. https://doi.org/10.1007/BF02834632

[29] Meta. 2023. *Llama 2: Open Foundation and Fine-Tuned Chat Models.* Meta Research Lab Technical Report. Meta.

[30] Meta. 2024. *The Llama 3 Herd of Models.* Meta Research Lab Technical Report. Meta.

[31] MistralAI. 2023. *Mistral 7B.* Mistral Research Lab Technical Report. MistralAI.

[32] Shunji Mori, Hirobumi Nishida, and Hiromitsu Yamada. 1999. *Optical character recognition.* John Wiley & Sons, Inc., USA.

[33] OpenAI. 2023. *GPT-4 Technical Report.* OpenAI Research Technical Report. OpenAI.

[34] Raymond Ptucha, Felipe Petroski Such, Suhas Pillai, Frank Brockler, Vatsala Singh, and Paul Hutkowski. 2019. Intelligent character recognition using fully convolutional neural networks. *Pattern Recognition* 88 (2019), 604–613. https://doi.org/10.1016/j.patcog.2018.12.017

[35] Alec Radford and Karthik Narasimhan. 2018. Improving Language Understanding by Generative Pre-Training. (2018). https://api.semanticscholar.org/CorpusID:49313245

[36] Connor Shorten and Taghi M. Khoshgoftaar. 2019. A survey on Image Data Augmentation for Deep Learning. *Journal of Big Data* 6, 1 (06 Jul 2019), 60. https://doi.org/10.1186/s40537-019-0197-0

[37] Shumaylov Z. Zhao Y. et al. Shumailov, I. 2024. AI models collapse when trained on recursively generated data. *Nature* 631 (2024), 755–759. https://doi.org/10.1038/s41586-024-07566-y

[38] Claus Smitt, Michael Halstead, Alireza Ahmadi, and Chris McCool. 2022. Explicitly Incorporating Spatial Information to Recurrent Networks for Agriculture. *IEEE Robotics and Automation Letters* 7, 4 (2022), 10017–10024. https://doi.org/10.1109/LRA.2022.3188105

[39] Yuan Y. Tang, Seong-Whan Lee, and Ching Y. Suen. 1996. Automatic document processing: A survey. *Pattern Recognition* 29, 12 (1996), 1931–1952. https://doi.org/10.1016/S0031-3203(96)00044-1

[40] Gemma Team. 2024. Gemma 2: Improving Open Language Models at a Practical Size. *ArXiv* abs/2408.00118 (2024). https://api.semanticscholar.org/CorpusID:270843326

[41] Maxim Tkachenko, Mikhail Malyuk, Andrey Holmanyuk, and Nikolai Liubimov. 2020-2022. Label Studio: Data labeling software. https://github.com/heartexlabs/label-studio Open source software available from https://github.com/heartexlabs/label-studio.

[42] Ashish Vaswani, Noam Shazeer, Niki Parmar, Jakob Uszkoreit, Llion Jones, Aidan Gomez, Łukasz Kaiser, and Illia Polosukhin. 2017. Attention is All you Need. In *Advances in Neural Information Processing Systems*, I. Guyon, U. Von Luxburg,

S. Bengio, H. Wallach, R. Fergus, S. Vishwanathan, and R. Garnett (Eds.), Vol. 30. Curran Associates, Inc. https://proceedings.neurips.cc/paper_files/paper/2017/file/3f5ee243547dee91fbd053c1c4a845aa-Paper.pdf

[43] Yiheng Xu, Minghao Li, Lei Cui, Shaohan Huang, Furu Wei, and Ming Zhou. 2020. LayoutLM: Pre-training of Text and Layout for Document Image Understanding. In *Proceedings of the 26th ACM SIGKDD International Conference on Knowledge Discovery amp; Data Mining (KDD '20)*. ACM. https://doi.org/10.1145/3394486.3403172

[44] Xu Zhong, Jianbin Tang, and Antonio Jimeno Yepes. 2019. PubLayNet: largest dataset ever for document layout analysis. In *2019 International Conference on Document Analysis and Recognition (ICDAR)*. IEEE, 1015–1022. https://doi.org/10.1109/ICDAR.2019.00166